\documentclass[]{spie}  

\usepackage{amsmath,amsfonts,amssymb}
\usepackage{graphicx}
\usepackage[colorlinks=true, allcolors=blue]{hyperref}
\usepackage{graphicx}
\usepackage{array}
\newcolumntype{M}[1]{>{\arraybackslash}m{#1}}
\usepackage{multirow,tabularx}
\usepackage{algpseudocode}

\title{Instructive artificial intelligence (AI) for human training, assistance, and explainability}

\author[a]{Nicholas Kantack}
\author[a]{Nina Cohen}
\author[a]{Nathan Bos}
\author[a]{Corey Lowman}
\author[a]{James Everett}
\author[a]{Timothy Endres}
\affil[a]{The Johns Hopkins Applied Physics Laboratory, Laurel, MD  20723}

\authorinfo{Send correspondence to nick.kantack@jhuapl.edu}

\begin{document} 
\maketitle

\begin{abstract}
We propose a novel approach to explainable AI (XAI) based on the concept of ``instruction" from neural networks. In this case study, we demonstrate how a superhuman neural network might instruct human trainees as an alternative to traditional approaches to XAI. Specifically, an AI examines human actions and calculates variations on the human strategy that lead to better performance. Experiments with a JHU/APL-developed AI player for the cooperative card game Hanabi suggest this technique makes unique contributions to explainability while improving human performance. One area of focus for Instructive AI is in the significant discrepancies that can arise between a human’s actual strategy and the strategy they profess to use. This inaccurate self-assessment presents a barrier for XAI, since explanations of an AI’s strategy may not be properly understood or implemented by human recipients. We have developed and are testing a novel, Instructive AI approach that estimates human strategy by observing human actions. With neural networks, this allows a direct calculation of the changes in weights needed to improve the human strategy to better emulate a more successful AI. Subjected to constraints (e.g. sparsity) these weight changes can be interpreted as recommended changes to human strategy (e.g. ``value A more, and value B less"). Instruction from AI such as this functions both to help humans perform better at tasks, but also to better understand, anticipate, and correct the actions of an AI. Results will be presented on AI instruction’s ability to improve human decision-making and human-AI teaming in Hanabi.
\end{abstract}

\keywords{XAI, explainability, interpretability, hanabi, human-machine, teaming, instructive, instruction}

\section{INTRODUCTION}

AI systems have demonstrated the ability to perform tasks remarkably well from games \cite{go, chess_and_shogi} to medical diagnosis \cite{breast_cancer}. Many of these AI are comprised of deep neural networks from which it is very hard to extract insight or explanations of the decision the network makes. Therefore, in many cases a neural network can discover novel insights into a domain but cannot communicate these insights to the humans that developed the network. This fundamental problem has sparked the active field of research into explainable AI (XAI)\cite{xai_survey}, AI for which there are some measures in place to facilitate human understanding of the AI's decision.

In some cases, the unexplainability of AI is a barrier to its use. Such cases are those in which humans are agents who must collaborate with the AI (which typically requires some level of common understanding) and those in which humans are significant stakeholders (e.g. when the AI is recommending medical treatment). A research effort at JHU/APL entitled ``Learning to Read Minds" studied this challenge in the context of human-machine teaming in the collaborative card game Hanabi \cite{l2rm}. Hanabi is a sort of cooperative solitaire with imperfect information that requires players (human or machine) to be able to infer the knowledge, intentions, and future actions from the behavior of their teammates \cite{hanabi_challenge}. Hanabi is a game for which the traditional process of self-play optimization (i.e. training an AI through millions of games played between copies of the same AI) does not lead to successful human-machine performance \cite{l2rm}, primarily because AI agents can develop obscure conventions (e.g. repurposing an in-game clue to mean something entirely different from its semantic meaning) that will be automatically understood by their mirror image during self-play, but completely incomprehensible to a human. This is why agents such as the Simplified Action Decoder \cite{sad}, Rainbow \cite{rainbow}, and Fireflower \cite{fireflower} \textit{often} achieve perfect scores in self-play, yet achieve low scores when playing with human teammates \cite{l2rm}. Furthermore, due to the lack of effective XAI techniques, there appear no practical means for these complex self-play AIs to explain these obscure conventions to humans (setting aside whether humans are even capable of implementing these conventions once understood).

\begin{figure}
\centering
\includegraphics[scale=0.8]{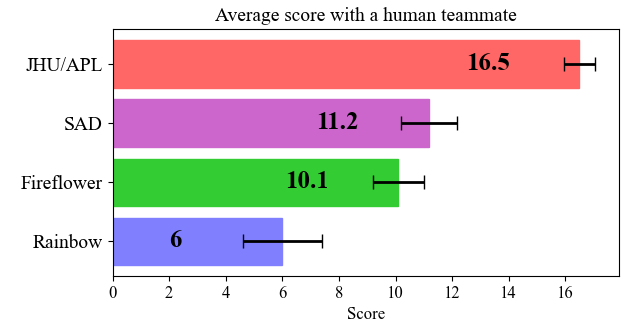}
\caption{The score distributions are shown for Simplified Action Decoder (a special off-belief version made for the competition), Rainbow, Fireflower, and the JHU/APL agent. These scores were obtained by pairing an agent with a human teammate (drawn from a pool of 21).}
\label{human_play}
\end{figure}

The ``Learning to Read Minds" research project included a JHU/APL-internal challenge tasking staff with developing AI agents that would excel when playing Hanabi with human strangers. The winning JHU/APL agent not only achieved human-play scores higher than any found in literature to date, \cite{other_play, intentional_hanabi, big_table} but it did so in a way that was constrained to allow human-readable descriptions of strategy (Figure \ref{human_play}). In particular, the JHU/APL agent demonstrated the ability to develop deep insights into human strategy through observation of human play, to understand how the human strategy interacted with the agent's strategy, and to adapt to discover a play style which complements the human strategy. This study summarizes the agent's structure which enabled it to successfully collaborate with human teammates, and introduces a novel type of explanation (we call ``instruction") to share AI insights with human observers.

\section{A Human-like Hanabi Agent}

The JHU/APL agent (henceforth referred to as ``agent") was developed under the philosophy that if it could play \textit{like} humans, it would play well \textit{with} humans. The agent was designed to convert the input space of the game state to a latent space of a small number of human-preferred factors (HPFs) which are aspects of the game that humans are known to attend to when making decisions. The agent utilizes twelve HPFs (Table \ref{factor_table}) which were suggested by intermediate Hanabi players. Constraining the attention of an AI in this fashion in order to guarantee some level of interpretability after training is a known practice \cite{explainability_constraint_1, explainability_constraint_2}. In the case of the JHU/APL Hanabi agent, an expected reward for each possible action is computed based on the expected effect the action will have on the HPFs. In particular, the expected value of an action is the inner product of a factor vector $\vec{h}$ with a weights vector $\vec{w}$. Therefore,
\begin{align}
y_{i} = \vec{h}_{i}^{t}\vec{w} \hspace{1cm}\implies \hspace{1cm} \vec{y}=H^{T}\vec{w}
\label{concise}
\end{align} 
where $y_{i}$ is the expected reward for action $i$, and vectors $\vec{h}_{i}$ form the columns of $H$. The elements of $\vec{h}$ are the expected changes that an action will induce on each of the HPFs (e.g. for the HPF of playing a playable card, the corresponding element in $\vec{h}_{i}$ is the probability that action $i$ will result in the playing of a playable card). The elements of $\vec{w}$ are the relative values of each HPF with respect to one another. Thus, while $H$ represents information about the game state, $\vec{w}$ represents the agent's \textit{strategy}. Altering the elements of $\vec{w}$ can dramatically alter the play style of the agent.

On each move, the agent calculates $\vec{y}$ which stores the expected reward for each possible action. The agent always chooses the action with the highest expected reward among the legal actions available. Of note, this technique does not involve and consideration of moves beyond the ply under consideration. Rather, the agent is pursuing an immediate improvement of the game state with respect to the chosen HPFs.

\begin{table}
  \caption{The twelve factors, along with their values for three different JHU/APL agent play styles (human-like, human-complementary, and self-play). To help interpret some of the factors, consider the following definitions. \textbf{endangered card} - a card for which no copy has been played, yet there is only one copy of this card remaining in play. \textbf{unneeded card} - A card which cannot be played in the future, for any reason.}
  \vspace{0.5em}
  \label{factor_table}
  \centering
  \begin{tabularx}{0.8\textwidth}{llll}
\hline
\multirow{2}{*}{Factor} & \multicolumn{3}{c}{Weights}\\ \cline{2-4}
& human-like & human-compl. & self-play\\
\hline\\
Playing a playable card & 1 & $\infty$ & 11\vspace{0.5em}\\
Playing unplayable card\\
(fewer than 2 strikes) & -1  & -1 & -1\vspace{0.5em}\\
Playing an unplayable card (2 strikes) & 3 & $\infty$ & 3\vspace{0.5em}\\
Other player playing a\\
playable card & 1.5 & 10 & 2\vspace{0.5em}\\
Other player playing an\\
unplayable card & 0 & 0 & 1\vspace{0.5em}\\
\hline\\
Discarding a non-endangered card & 0.1 & 0.55 & 0.8\vspace{0.5em}\\
Discarding an unneeded card & 0.25 & 1 & 0\vspace{0.5em}\\
\hline\\
Playing a singled out card & 3 & 1.5 & 5\vspace{0.5em}\\
Giving a clue that singles\\
out a playable card & 3 & 3 & 2\vspace{0.5em}\\
Giving a clue that singles\\
out a non-playable card & 0 & -5 & 4\vspace{0.5em}\\
Discarding a singled out card & -0.5 & -2 & -3\vspace{0.5em}\\
Added value to any clue\\
per info token held & 0.5 & 0.1 & 0\vspace{0.5em}\\
\hline
\end{tabularx}
\end{table}

\subsection{Modeling Human Decision Making}

\begin{figure}
\centering
\includegraphics[scale=0.8]{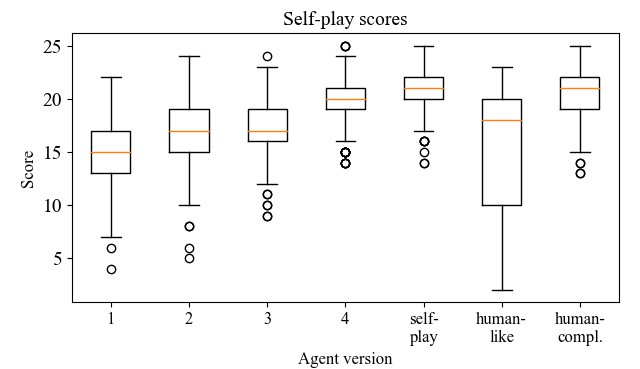}
\caption{Self play scores are shown for different versions of the agent during the initial development process. Each transition from one version to the next was accompanied by changes to the weights vector $\vec{w}$ or the addition of new elements to the weights vector. Notably, the human-like agent had significantly poorer self-play scores, consistent with the fact that the human-like agent was optimized to agree with a database of human decisions.}
\label{self_play}
\end{figure}

While human-like play was the preliminary goal during the agent's development, the first training efforts were aimed at generating decent self-play scores. For this phase, the training of the agent was separated into epochs. Each epoch consists of a four-dimensional, full factorial design experiment on a subset of four elements from $\vec{w}$. Each element under test was given three test values (a low, medium, and high) around the neighborhood of where the optimal value was expected to be. Therefore, each epoch tested $3^{4}=81$ unique $\vec{w}$ vectors. For each $\vec{w}$ vector tested, 200 games were played between identical copies of the agent. The elements of $\vec{w}$ under test were not altered until a an epoch occurred for which the highest score was achieved by assigning the medium value for each element under test (i.e. increasing or decreasing any element led to poorer performance). The progression of self-play scores during this development phase are shown in Figure \ref{self_play}.

\begin{figure}
\centering
\includegraphics[scale=0.8]{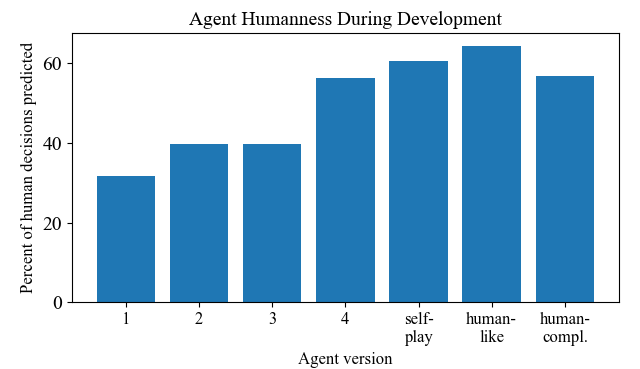}
\caption{The humanness of major versions of the agent are shown. Humanness is defined as the fraction of human decisions with which the agent agrees when analyzing a database of 376 game decisions made by one of the authors. Of the agents shown, only the ``human-like" agent was explicitly optimized for maximal humanness. The considerable humanness of the other models indicates how optimizing an HPF focused agent for self play can lead to considerably human-like performance.}
\label{humanness}
\end{figure}

Once the agent was optimized for self-play, the next objective was to find a strategy vector $\vec{w}$ that would lead to play that was as human-like as possible. To facilitate this exploration, a dataset of 376 decisions was collected by examining the play of one of the authors. With a dataset of decision made by a single human, we aimed to determine if a strategy vector $\vec{w}$ could be fitted to a particular human's play style rather than a $\vec{w}$ which represented some ambiguous (perhaps bad) play style that was averaged across humans with potentially dissimilar play styles. The progression of increasing humanness is displayed in Figure \ref{humanness}. The highest humanness fraction of any agent was 64.2\%, achieved by the human-like agent (that is, the agent was able to independently agree with the human decision in 64.2\% of the game states examined).

\begin{figure}
\centering
\includegraphics[scale=0.8]{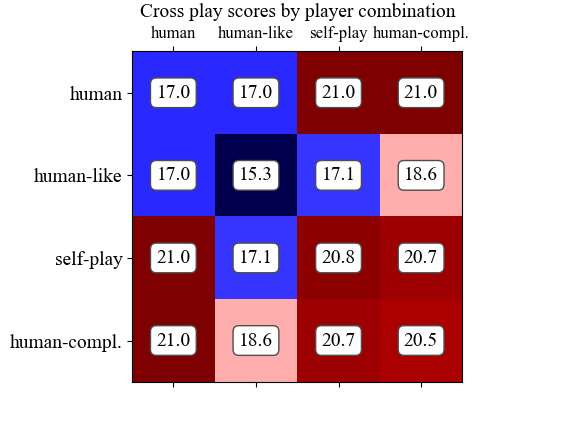}
\caption{Average scores are shown for different pairings of the agents (and humans). The human-human score is included for reference, but is an average across a small number of games played within the development team. All other scores are averages across at least 10 games. Error margins for each bin are less than 1 point. It is clear that the human-complementary agent achieves the highest score of any agent paired with the human-like agent (since these are the exact conditions under which the human-complementary agent was optimized). It is notable that the average human + self-play and human + human-compl. scores are identical in this plot (where a difference is expected), but it is also worth noting that a very small number (two) of very experienced humans were represented in these data. Therefore, while these scores are useful for comparing how non-human agents perform in different pairings, generalizations about human play from this figure should be made with great caution.}
\label{cross_play}
\end{figure}

Once the human-like version of the agent had been optimized for fitting the dataset of human decisions, a final training effort was made by pairing a training version of the agent with the human-like version. In this fashion, the training process was intended to approximate playing with a human teammate. As before, full factorial design experiments were run altering four elements of $\vec{w}$ per epoch, each across three levels. At then end of the training process, the ``human-complementary" version of the agent was created. Cross play results (Figure \ref{cross_play}) illustrate the performance of different combinations of agents developed.

\subsection{An Important Note on Human Perception of Strategy}

It is worth noting that the JHU/APL agent needed to make significant changes to its initial $\vec{w}$ vector in order to accurately predict human decision making in the game, despite the fact that the initial $\vec{w}$ given to the agent was intended to accurately describe human decision making. For this reason, it became clear that human players could not accurately depict the weights they attributed to HPFs. This has profound implications for XAI. A common XAI approach would have involved taking the values from the self-play strategy vector in Table \ref{factor_table} and describing these to a human player (e.g. ``You should value discarding a non-endangered card at 0.8"). However, if a human already egregiously misunderstands what value they actually attribute to these HPFs, it is unlikely that the human will be able to act on this insight. Rather, it would perhaps be more suitable for us to look at the difference in weights between the human-like agent and the self-play agent, since doing so would allow us to specify corrections a human should make to their strategy (e.g. ``you should value discarding a non-endangered card more"). These corrections are human interpretable \textit{regardless} of whether the human accurately understands their current strategy. This is the principal idea behind AI instruction.

\section{Theory of AI Instruction}

AI instruction is defined in the context of explaining differences in strategy in the form of changes on weights. Therefore, it is relevant to consider a difference in outputs (say, $\vec{y}_{1}$ from $\vec{w}_{1}$ and $\vec{y}_{2}$ from $\vec{w}_{2}$).
\begin{align}
\delta \vec{y} = \vec{y}_{1} - \vec{y}_{2} = H^{T}\vec{w}_{1} - H^{T}\vec{w}_{2}
\end{align}
\begin{align}
\delta \vec{y} = H^{T}\left(\vec{w}_{1} - \vec{w}_{2}\right) = H^{T} \delta \vec{w}
\label{strat_difference}
\end{align}

\subsection{A note on Strategy vs. Perception}

It is possible to imagine a difference in performance, $\delta \vec{y}$, arising not from a difference in strategy ($\delta \vec{w}$), but rather a difference in perception of the game state $\delta H$. This is particularly practical if the elements of $H$ concern complex changes in the game state such as the probabilities of certain outcomes (as it does in Hanabi). In this case,
\begin{align}
\delta \vec{y} = \delta H^{T}\vec{w}
\end{align}
In fact, there is ambiguity between this and \eqref{strat_difference}, since $HH^{T}$ is a $12 \times 12$ matrix that will tend to be full rank, and thus the relation
\begin{align}
\delta \vec{w} = \left(HH^{T}\right)^{-1}H\delta H^{T} \vec{w}
\end{align}
indicates that any strategic difference $\delta \vec{w}$ could be interpreted instead as an observation error $\delta H$. This illustrates yet another reason for providing instruction in the form of $\delta \vec{w}$ rather than an explanation in the form of $\vec{w}_{2}$. By the very nature of this equivalence relation, a recommended change in strategy, $\delta \vec{w}$ can compensate for both misperception and strategic deficiency (and mixtures thereof).

\subsection{Non-uniqueness of and Constraints on $\delta\vec{w}$}
Since $H^{T}$ is a tall matrix ($20 \times 12$), $H^{T}$ has a non-empty null space. Therefore, the condition
\begin{align}
\delta \vec{y} = H^{T}\left( \delta \vec{w} + \vec{n} \right)
\end{align}
is satisfied for any $\vec{n}$ in the null space of $H^{T}$, and so any $\delta \vec{w} + \vec{n}$ is a valid description of a strategic difference needed to elicit the decision difference $\delta \vec{y}$. This non-uniqueness of $\delta \vec{w}$ is advantageous, because it allows multiple possible $\delta \vec{w}$ to be compared for fitness according to human friendly constraints (e.g. norm minimality, sparsity, etc.).

\subsection{Generating AI Instruction}
Suppose that a human subject is presented with $g$ game states $H_{1},...,H_{g}$, and that these game state matrices are stacked into a $12 \times 20 \times g$ tensor $\mathcal{H}$. As before, a strategy $\vec{w}$ can be combined with a game state to yield a vector of outputs, $\vec{y}$.
\begin{align}
\vec{y}_{k} = \mathcal{H}(:, :, k)^{T}\vec{w}
\end{align}
In the following formalism, a subscript $h$ corresponds to a human, while a subscript $i$ corresponds to an ideal (typically a successful AI). Let's assume that the index of the maximum element of $\vec{y}_{h}$ indicates the decision that will be taken (per strategy $\vec{w}_{h}$) for the game state used. In this case, two different $\vec{y}$ vectors may still specify the same action if their maximal elements occupy the same index. If not, then it is worth describing the nearest (min $|\vec{y}_{i} - \vec{z}|$) vector $\vec{z}$ such that $\vec{y}_{h}$ and $\vec{z}$ have maximal elements in the same index position (a position different than the maximal element of $\vec{y}_{h}$). Suppose $y_{h}(t)=$max$(\vec{y}_{h})$, but no other information is known about $\vec{y}_{h}$. Then let $m$ be the average of all the terms of $\vec{y}_{i}$ that are greater than $y_{i}(t)$. Then $\vec{z}$ is defined as
\begin{align}
z(i) = \left\lbrace\begin{matrix}
y_{i}(j) & y_{i}(j) < m\ \& \ i \neq t \\
m + \varepsilon & y_{i}(j) < m\ \& \ j = t \\
m & y_{i}(j) > m \\
\end{matrix}
\right.
\label{z_calc}
\end{align}
where $\varepsilon$ is some small, positive tie-breaking factor. If the vectors $\vec{z}$ (of which there are $g$) are made the columns of a $20 \times g$ matrix $Z$, and the output vectors $y_{i}$ from strategy $\vec{w}_{i}$ are made columns of a $20 \times g$ matrix $Y_{i}$, then we can relate $Z$ and the game state tensor $\mathcal{H}$ as follows.
\begin{align}
\mathcal{H}(:, :, k)^{T} \delta \vec{w} = Z(:, k) - Y_{i}(:, k) \ \forall \ k \in [1, g]
\end{align}
Where $\delta \vec{w}$ is a strategy change needed so that $\vec{w}_{i} + \delta \vec{w}$ and $\vec{w}_{h}$ arrive at the same decision for every game state in $\mathcal{H}$. To calculate for $\delta \vec{w}$, we can utilize the following matrix unfolding.
\begin{align}
\left[
\begin{matrix}
\mathcal{H}(:, :, 1)^{T} \\
\mathcal{H}(:, :, 2)^{T} \\
... \\
\mathcal{H}(:, :, g)^{T}
\end{matrix}
\right] \delta \vec{w} = \widetilde{H} \delta \vec{w} = \text{vec}\left(Z - Y_{i}\right)
\label{linear_system}
\end{align}
This is an overspecified linear system, so a least squared error solution can be taken for $\delta \vec{w}$. Then, $\delta \vec{w}$ is the norm-minimal change to apply to $\vec{w}_{i}$ to better concur with strategy $\vec{w}_{h}$ in each of the $g$ game states. If $\vec{w}_{h}$ is the strategy of a human, and $\vec{w}_{i}$ is an ideal, $\delta \vec{w}$ is the change \textit{to the ideal} needed to concur with the human. The inverse ($-\delta \vec{w}$) has elements which comprise the instructions that should be given to the human. In essence, the instructed changes are the opposite of those needed for the ideal to be altered to make the same decisions the human made.

\subsection{Properties of the Generated $\delta \vec{w}$}
$\delta \vec{w}$ is not guaranteed to produce consensus between the starting strategy and the ideal when adopted. Formally, it does not always hold that
\begin{align}
\mathcal{H}(:, :, k)(\vec{w}_{i} + \delta \vec{w}) = \mathcal{H}(:, :, k)\vec{w}_{h}
\end{align}
However, it is possible (and desired) for this relation to circumstantially hold for many $k$ values. Increasing $\varepsilon$ in the above formulation will tend to increase the number of game states in which consensus is built but at the expense of a larger norm $\delta {w}$ (i.e. bigger recommended changes to $\vec{w}_{h}$). Even so, total consensus between the ideal and modified strategies is rarely achieved because the model for decisions generated from game states given by \eqref{concise} may not accurately describe all decisions ($\vec{y}$) made by a human (e.g. due to momentary misperception, distraction, and attention to factors not captured in $H$). The quantity $\lambda$ defined as
\begin{align}
\lambda = \min_{\vec{w}}\frac{1}{g}\sum_{k=1}^{g}|\vec{y}_{h} - \mathcal{H}(:, :, k)^{T}\vec{w}|
\end{align}
may be introduced as a figure of merit for the list of factors which define the strategy vector $\vec{w}$. Furthermore, $\lambda$ can be used to measure the utility of elements of $\vec{w}$ by examining the change in $\lambda$ induced by the removal or inclusion of factors. Ideally, the only factors kept would be those whose inclusion result in a significant decrease in $\lambda$.
\par Similarly, one can define a figure of merit for generated instruction. If we define $f(\vec{a}, \vec{b})$ as
\begin{align}
f(\vec{a}, j)=
\left\lbrace
\begin{matrix}
1 & a(j) = \text{max}(\vec{a}) \\
0 & \text{otherwise}
\end{matrix}
\right.
\end{align}
If $n_{k}$ is the index of the decision the human instructee made for game state $k$, then it is possible to evaluate the quality $q(\delta \vec{w})$ of instructions as
\begin{align}
q(\delta \vec{w}) = \frac{1}{g} \sum_{k=1}^{g}f(\mathcal{H}(:, :, k)^{T}(\vec{w}_{i} + \delta \vec{w}), n_{k})
\end{align}
$q(\delta \vec{w})$ falls in $[0, 1]$ and can be interpreted as the fraction of human decisions that can be understood as a variation ($\delta \vec{w}$) on an ideal ($\vec{w}_{i}$).

\subsection{Full AI Instruction Algorithm with Quality Monitoring}

We recommend the algorithm in Figure \ref{iai_algorithm} for generating AI instruction. The algorithm has two preparation steps. The first is to train up an ideal strategy ($\vec{w}_{i}$), and the second is to aggregate a dataset of human decisions $\vec{n}$ paired with the game states in which they were made (slabs of the $\mathcal{H}$ tensor). While it is possible to terminate the algorithm after the step that assigns $\delta \vec{w}$, this algorithm includes a post-processing component which seeks to zero out as many elements of $\delta \vec{w}$ as possible while maintaining some preset explanatory fidelity $\alpha$ to the human decision set. The purpose of this post-processing is to generate instruction which concerns changes in as few of values as possible. This is motivated by the assumption that low dimensional instructions are easier for humans to understand (i.e. require focusing on fewer aspects of the game in subsequent play).

\begin{figure}
\textbf{Instructive AI Algorithm}
\hrule
\begin{algorithmic}
\State {$\alpha \gets \text{Chose accuracy threshold in [0,1]}$}
\State {$\vec{w}_{i} \gets \text{Train}(\hat{\mathcal{H}}, C)$ Learn ideal weights on a dataset $\hat{\mathcal{H}}$}
\State {$\mathcal{H},\vec{n} \gets \text{Observe }g\text{ human decisions}$}
\For{$k \in [1, g]$}
\State $Y_{i}(:, k) \gets \mathcal{H}(:, :, k)^{T}\vec{w}_{i}$
\State $Z(:,k) \gets (Y_{i}(k), n(k))$ per \eqref{z_calc}
\EndFor
\State {$\delta \vec{w} \gets \text{Solve }\widetilde{H}\delta\vec{w}=\text{vec}\left(Y_{i} - Z\right)$}
\State {$q \gets \frac{1}{g}\sum_{k=1}^{g}f(\mathcal{H}(:, :, k)^{T}(\delta \vec{w}_{i} + \delta \vec{w}), n_{k})$}
\While {$q>\alpha$}
\State {$\delta \vec{w} \gets $ zero out element with smallest impact on $q$}
\State {$q \gets \frac{1}{g}\sum_{k=1}^{g}f(\mathcal{H}(:, :, k)^{T}(\delta \vec{w}_{i} + \delta \vec{w}), n_{k})$}
\EndWhile
\State Give $-\delta \vec{w}$ to human
\end{algorithmic}
\caption{This algorithm generates sparse AI instruction tailored to a set of human decisions.}\label{iai_algorithm}
\end{figure}

\section{Experimental Results}

\begin{figure}
\centering
\includegraphics[scale=0.6]{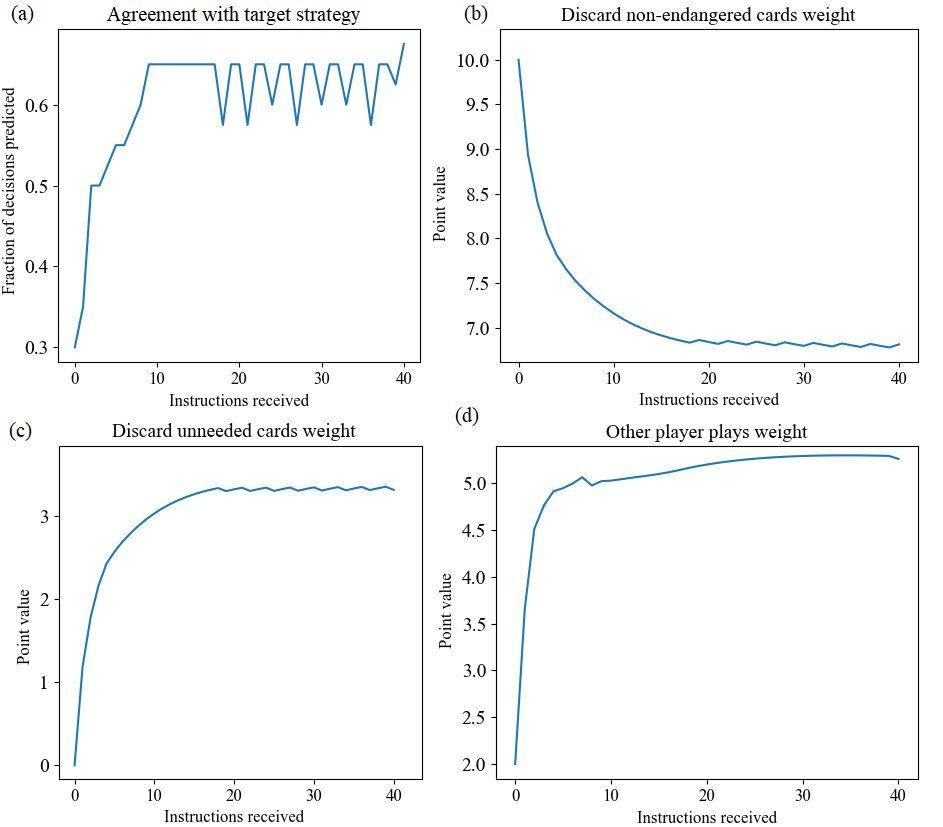}
\caption{The effect of following successive batches of AI instructions are shown. For this experiment, a trial version of the self-play agent was created that had a non-endangered discard weight of 10 rather than the normal 0.8. Instructions are generated for how this agent can better emulate the self-play agent. (a) The agreement between the agents is shown as a function of how many batches of instructions were generated. After each batch, the trial agent applies the correction to its weights and reexamines agreement. (b) The trial agent's inflated discard weight decreases with time. Notably, it does not appear to stabilize at the same value as the ideal (0.8). (c) The weight for discarding unneeded cards is shown over the experiment. Note that while the two agents initially had the same value (0), this weight drifts to a new value as a result of the instructions, indicating that a new set of weights is being discovered that agrees with the self-play agent over the decisions studied. (d) Another weight is shown (which does not pertain to discards) to further illustrate how instruction may not encourage the same, unique weights as the ideal. }
\label{instructions}
\end{figure}

During the ``Learning to Read Minds" challenge, a database of 376 human decisions in Hanabi games was generated. Preliminary results are shown based on analysis of this dataset. To illustrate the instruction generation process, a trial agent was created by copying the self-play agent. Because the self-play agent already agrees with human decision at a high rate, the weight for the non-endangered discard was inflated (to a value of 10). Then, in an iterative process, instructions were generated (on how the trial agent could better emulate human decision making based on the dataset), the trial agent applied the instructed changes to its weights, and a new set of instructions were generated. This process is shown to lead to asymptotic improvement in the agreement between the trial agent and the human dataset (Figure \ref{instructions}).

High agreement ($68\%$) was achievable after 40 instruction based weight updates. Furthermore, the spurious discard weight was shown to be brought into closer agreement with the target strategy. Importantly, this (and other initially matching weights) were shown to drift to new equilibrium values. This serves as an empirical demonstration of the non-uniqueness of strategies as described in the previous section. However, it is important to note that the generation of a norm minimally different $Z$ matrix may not provide a linear system in \eqref{linear_system} that admits a solution that produces high prediction accuracy when observing the target strategy. This is because the matrix $Z$ may be a poor estimation for the target strategy's output vectors, a circumstance that is increasingly likely when the instructee strategy differs significantly from the target strategy.

These results (Figure \ref{instructions}) indicate that AI instruction can indeed provide stepwise improvements to strategy which, taken iteratively, can lead to significant improvement in the agreement between the instructee strategy and the ideal. In this way, instructions serve as something of a proxy gradient of a cost function, namely, agreement with the ideal. Utilizing the instructions as a gradient for agent training was shown in this experiment to lead to better humanness scores ($68\%$ vs. $64\%$) in a much shorter computation time (minutes vs. hours) compared to the full factorial approach. Additionally, these instructions provide a novel approach to portraying AI insight in a way that is understandable to human observers. Specifically, these instructions can be phrased as corrections to the weights attributed to human-preferred factors, allowing for AI systems to develop an understanding of human decision making and to share those insights through tailored instructions.

\section{Conclusion}

Leveraging insights obtained from the development of a highly successful, artificially intelligent human teammate for Hanabi, we propose a technique of instructive AI to better enable humans to obtain insight from complicated AI systems. There are assumptions in this approach that may not hold true for certain contexts. For instance, this technique hopes that the requisite $\delta \vec{w}$ for consensus building is small. If not, then implementing a $\delta \vec{w}$ may be just as confusing for humans as being told $\vec{w}_{i}$, or perhaps even more so. More fundamental, the model given by \eqref{concise} may not accurately capture a majority of a human's decisions, and $\vec{w}$ is always at risk of missing elements that are crucial to a human's decision making. In general, it is challenging to produce a complete set of values relevant to human decision making. For the purposes of this experiment, the list of values is produced from human introspection and trial and error. Techniques to organically learn the needed values may be possible and highly valuable to the task of generating AI instruction, but are beyond the scope of the experiments described above.
\par Many of the challenges described above apply in similar form to other methods of XAI. However, instructive AI shows promise to circumvent some of the greatest challenges of XAI and provide a novel framework in which further research might push the frontier on extracting human-useful insight from complex AI systems.

\bibliography{ref} 
\bibliographystyle{spiebib} 

\end{document}